\pdfoutput=1
\documentclass[11pt]{article}

\usepackage{acl}
\usepackage{times}
\usepackage{latexsym}
\usepackage{float}

\usepackage[T1]{fontenc}
\usepackage[utf8]{inputenc}

\usepackage{microtype}
\usepackage{booktabs}
\usepackage{multicol,multirow}

\usepackage{inconsolata}

\usepackage{graphicx}
\usepackage{amsfonts,amsmath,amssymb,amsthm}
\usepackage[ruled,vlined]{algorithm2e}
\usepackage{algorithmic}

\makeatletter
\renewcommand{\@fnsymbol}[1]{$\dagger$}
\makeatother

\title{WPO: Enhancing RLHF with Weighted Preference Optimization}

\author{Wenxuan Zhou\thanks{Correspondence to <\texttt{wenxuan.zhou@zoom.us}>}, Ravi Agrawal, Shujian Zhang, Sathish Reddy Indurthi\\
\textbf{Sanqiang Zhao, Kaiqiang Song, Silei Xu, Chenguang Zhu}\\
Zoom Video Communications}

\begin{document}
\maketitle
\begin{abstract}
Reinforcement learning from human feedback (RLHF) is a promising solution to align large language models (LLMs) more closely with human values.
Off-policy preference optimization, where the preference data is obtained from other models, is widely adopted due to its cost efficiency and scalability.
However, off-policy preference optimization often suffers from a distributional gap between the policy used for data collection and the target policy, leading to suboptimal optimization.
In this paper, we propose a novel strategy to mitigate this problem by simulating on-policy learning with off-policy preference data.
Our \underline{W}eighted \underline{P}reference \underline{O}ptimization (WPO) method adapts off-policy data to resemble on-policy data more closely by reweighting preference pairs according to their probability under the current policy.
This method not only addresses the distributional gap problem but also enhances the optimization process without incurring additional costs.
We validate our method on instruction following benchmarks including Alpaca Eval 2 and MT-bench.
WPO not only outperforms Direct Preference Optimization (DPO) by up to 5.6\% on Alpaca Eval 2 but also establishes a remarkable length-controlled winning rate against GPT-4-turbo of 76.7\% based on Gemma-2-9b-it.
We release the code and models at \url{https://github.com/wzhouad/WPO}.
\end{abstract}

\section{Introduction}
Large language models (LLMs; \citealt{ouyang2022training,achiam2023gpt,tunstall2023zephyr,chung2024scaling}) have demonstrated remarkable capabilities in generating human-like responses. However, they still face challenges in scenarios demanding high standards of reliability, safety, and ethics.
To address these challenges, reinforcement learning from human feedback (RLHF; \citealt{christiano2017deep, ouyang2022training, glaese2022improving}) is a promising approach to better align LLMs with human values.

Depending on how the outputs are generated, RLHF can be categorized into on-policy and off-policy settings.
In the on-policy setting~\cite{schulman2017proximal,yuan2024self,rosset2024direct,wu2024self}, the policy model used to generate outputs is the same as the policy model being optimized.
During this process, a policy model is first initialized from supervised finetuning (SFT).
Then, a reward model~\cite{schulman2017proximal,gao2023scaling,jiang-etal-2023-llm} is obtained based on human~\cite{schulman2017proximal} or AI~\cite{lee2023rlaif} feedback.
Finally, the policy model samples outputs during training, which are then evaluated using the reward model.
The policy model is optimized to improve the expected reward using training objectives such as Proximal Policy Optimization (PPO; \citealt{schulman2017proximal}) and Direct Preference Optimization (DPO; \citealt{rafailov2024direct}).
However, on-policy RL relies heavily on policy sampling during training and online rewards, which can incur high costs.
In contrast, in the off-policy setting~\cite{tunstall2023zephyr,ivison2023camels}, the outputs are generated from different models, and the policy model is optimized based on these data instead of its sampled outputs.
Consequently, off-policy RL offers significant advantages in terms of cost and data efficiency and is easier to scale up.%is more scalable in scenarios where collecting new outputs and rewards is expensive or impractical.

\begin{figure*}[t]
\centering
\includegraphics[width=16cm]{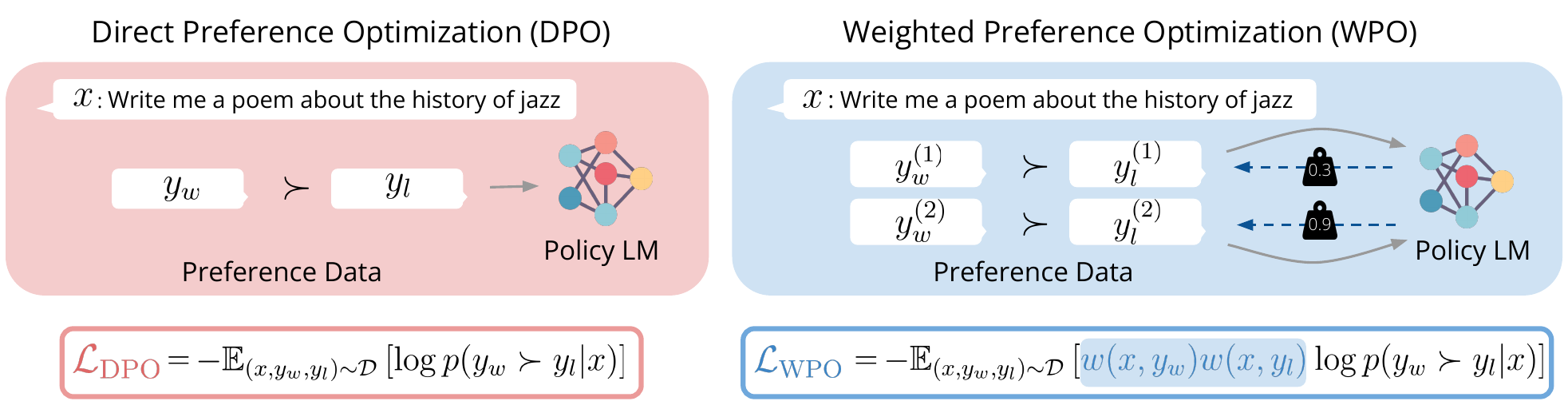}
\caption{Overview of the Weighted Preference Optimization (WPO). Some notations are labeled along with corresponding components. Existing DPO directly optimizes the policy to best satisfy the preferences with off-policy data. In contrast, WPO adapts off-policy data to resemble on-policy data more closely by reweighting preference pairs according to their probability under the current policy. 
}
\label{fig:pipeline}
\vspace{-10pt}
\end{figure*}

Nevertheless, off-policy RL often shows worse performance than on-policy RL, due to the distributional gap between the policy used to collect data and the target policy being optimized, which leads to instability and inefficiency in training~\cite{fujimoto2019off,kumar2019stabilizing,kumar2020conservative,xu2024dpo,tang2024understanding,tajwar2024preference}.
In off-policy preference optimization, the optimization is typically performed on preference data sampled from other models, and all the preference singles are equally treated.
%However, when using LLMs, we typically sample from the most probable outputs.
% \cz{too strong assumption. Also why?}
However, some preference data, distant from the current policy, are less informative for training, resulting in inefficient and suboptimal optimization.
%Making preference optimization on-policy~\cite{yuan2024self,rosset2024direct} can mitigate this problem, yet it requires additional training costs and data acquisition efforts.
%Moreover, within these methods, new preference data is sampled per iteration rather than per optimization step, where distribution shift can still occur between iterations.

In this paper, we propose simulating on-policy preference optimization with off-policy preference data, combining the efficiency of off-policy RL with the performance benefits associated with on-policy RL.
Our method is motivated by the following conceptual data generation process.
This process begins with transforming the existing preference dataset into a preference labeling function.
We can then resample a new preference dataset through bootstrapping from the existing data.
This process involves uniformly sampling inputs from the preference dataset and online sampling new pairs of outputs with the current policy model.
Each pair is retained if it can be labeled by the labeling function; otherwise, it is rejected.
We then perform DPO on the regenerated preference dataset.
In practice, this bootstrapping process can be implemented with the \underline{W}eighted \underline{P}olicy \underline{O}ptimization (WPO) objective, where different preference pairs are reweighted according to the joint probability of their outputs.
We further devise a weighting alignment mechanism to ensure that all on-policy generated pairs are equally weighted.
In this way, WPO can effectively mitigate the distribution gap during RL without incurring additional costs.

We evaluate WPO on instruction following benchmarks, including Alpaca Eval 2~\cite{alpaca_eval} and MT-bench~\cite{zheng2024judging}.
In the off-policy setting based on Ultrafeedback~\cite{cui2023ultrafeedback}, WPO improves the length-controlled winning rate against GPT-4-turbo on Alpaca Eval 2 by up to 14.9\% over SFT model, outperforming DPO by up to 5.6\%.
Particularly, in the hybrid RL setting where the off-policy preference data is further enriched with on-policy outputs, WPO (Figure \ref{fig:pipeline}) achieves a new SOTA length-controlled winning rate of 76.7\% on Alpaca Eval 2.
Additionally, we find that WPO can be integrated into other loss functions for preference optimization and shows consistent improvements.
Furthermore, we systematically compare the model performance in different RL settings. Our analysis reveals that the hybrid setting, which utilizes both on-policy and off-policy preference data, achieves the best results, and on-policy, dispreferred data is more important for preference optimization.

To summarize, our contributions are three-fold:
\begin{itemize}
\setlength{\itemsep}{0pt}
\setlength{\parsep}{0pt}
\setlength{\parskip}{0pt}
\item We identify the distribution gap problem in off-policy preference optimization, and accordingly introduce a method to simulate on-policy RL using off-policy preference data.
\item We propose the WPO objective, which reweights preference pairs based on their probabilities.
This ensures that the most relevant and probable outputs are prioritized during optimization, mitigating the distribution gap and improving the effectiveness of the preference optimization.
\item We conduct extensive instruction following benchmarks.
Our results demonstrate that WPO significantly outperforms DPO and achieves new SOTA results on Alpaca Eval 2 in the hybrid RL setting.
\end{itemize}

\section{Related Work}
\paragraph{General alignment methods.}
The advancement of ChatGPT has propelled significant advancements in the field of large language models (LLMs). Notable models such as Zephyr \cite{tunstall2023zephyr} and GPT-4 \cite{achiam2023gpt} have effectively demonstrated the application of techniques like reinforcement learning from human feedback (RLHF; \citealt{christiano2017deep, ouyang2022training, glaese2022improving}) and direct preference optimization (DPO; \citealt{rafailov2024direct}), highlighting their efficacy in achieving improved model alignment. These approaches, along with related methods such as sequence likelihood calibration \cite{zhao2023slic} and Generalized Preference Optimization (GPO) \cite{tang2024generalized}, aim to refine the objectives of RLHF by clearly enhancing the distinction between more and less preferred outputs. Additionally, the introduction of the Direct Nash Optimization (DNO) algorithm by \citet{rosset2024direct} represents a further innovation. 
This algorithm utilizes cross-entropy to assess the gap between actual and predicted win rates. 
Practical applications more frequently rely on the iterative framework of DPO~\cite{xu2023some}. Yet, DPO often reveals a discrepancy between the output distributions produced by the policy and those in the preference dataset. To address this, we propose simulating on-policy reinforcement learning using off-policy data, thereby combining the benefits of on-policy RL with enhanced efficiency.

\paragraph{On-policy reinforcement learning.}
Self-Play Fine-Tuning \cite{chen2024self} operates under an iterative framework akin to DPO, utilizing human-labeled responses as "winners" and outputs from previous iterations as "losers" within each pairing. Similarly, Adversarial Preference Optimization \cite{cheng2023adversarial} incorporates contrastive losses, which obviate the need for direct feedback from annotators. This method introduces a token-level loss function known as Cringe Loss \cite{adolphs2022cringe}, which differentiates the correct subsequent token from a deliberately incorrect token from the vocabulary. Pairwise Cringe Loss \cite{xu2023some} utilizes this cringe loss mechanism within a continuously improving iterative training framework. Moreover, the recent introduction of SAMI \cite{franken2024self} targets optimizing a lower bound on the conditional mutual information between prompts and responses through a contrastive estimation technique. In our approach, we adjust the importance of each pair in the training process by assigning greater weight to those pairs more likely to be sampled from the policy model, thus simulating on-policy reinforcement learning.

\section{Method}
In this section, we provide the theoretical background of RLHF and DPO in Section~\ref{ssec:preliminary}.
We then introduce the distributional gap problem and propose the WPO method (Algorithm \ref{alg:wpo}) in Section~\ref{ssec:wpo}.
Finally, we explore how to better simulate on-policy RL through weight alignment in Section~\ref{ssec:weight_alignment}.

\begin{algorithm}[t]
\caption{Weighted Preference Optimization (WPO)} 
\label{alg:wpo}
\begin{algorithmic}
\STATE \textbf{Input:} Dataset ($\mathcal{D}$) with prompts and respon-\\ses, policy  LM $\pi_{\theta}$, total number of iterations \\$T$, learning rate $\alpha_t$,
%denotes the time step. 
\FOR{$t =0$  to $T$}
% \FOR{each win response $y_{w_i}$ and lose response $y_{l_j}$ in the batch}
\STATE Sample a mini-batch of tuples $(x, y_w, y_l)$ from $\mathcal{D}$,
\STATE Calculate the alignment weight via Eq. \eqref{eq:weight_sample},
\STATE Compute  $\mathcal{L}_\text{WPO}$ via Eq. \eqref{eq:wpo_loss}, 
% \IF{ $t \% u == u - 1$}
\STATE Update policy parameters $\theta$ using gradient descent: $\theta  \leftarrow \theta  -  \alpha_t\nabla \theta(x, y_w, y_l, \theta)$.
% \ENDIF
 \ENDFOR
 % \ENDFOR
\end{algorithmic}
\vspace{-2pt}
\end{algorithm}

\subsection{Preliminaries}
\label{ssec:preliminary}
RLHF~\cite{schulman2017proximal} aims to align a large language model with human preferences.
Given a preference dataset $\mathcal{D}=\{(x^{(i)}, y_w^{(i)}, y_l^{(i)})\}_{i=1}^N$, in which $y_w$ and $y_l$ are a pair of outputs given prompt $x$ sampled from a policy model, and $y_w$ is favored over $y_l$ as determined by human or AI annotators.
This preference is modeled by a latent reward function $r^*(x, y)$, which scores on how well the candidate output $y$ matches the input $x$.
There are various ways to model the reward function, among which the Bradley-Terry (BT; \citealt{bradley1952rank}) model is most commonly used.
The BT model assumes that the preference distribution is characterized by the following equation:
\begin{equation*}
\resizebox{0.46\textwidth}{!}{
    $p(y_w \succ y_l|x)=\frac{\exp (r^*(x, y_w))}{\exp (r^*(x, y_w))+\exp(r^*(x, y_l))}$}.
\end{equation*}
The parameters of the reward function can be estimated based on maximum likelihood estimation, resulting in the reward model $\hat{r}(x,y)$.
Then, we can use the fitted reward model to provide feedback to a large language model by optimizing the following objective:
\begin{equation*}
\resizebox{0.46\textwidth}{!}{
    $\max_{\pi_\theta} \mathbb{E}_{x\sim \mathcal{D},y\sim \pi_\theta(\cdot|x)}\left[\hat{r}(x,y)-\beta\log\frac{\pi_\theta(\cdot|x)}{\pi_\text{ref}(\cdot|x)}\right]$,}
\end{equation*}
where $\beta$ controls the deviation between the policy model $\pi_\theta$ and the reference model $\pi_\text{ref}$, which is usually initialized from the SFT model.

\smallskip
\noindent \textbf{DPO.}
Direct optimization optimization (DPO; \citealt{rafailov2024direct}) integrates the learning of the reward function and the policy model to a unified objective.
Specifically, suppose the optimal policy $\pi^*$ is given, the corresponding reward $r^*$ has a closed form:
\begin{equation*}
    r^*(x,y)=\beta \log\frac{\pi^*(y|x)}{\pi_\text{ref} (y|x)} + \beta \log Z(x),
\end{equation*}
where $Z(x)$ is the partition function.
Applying this reparameterization to the BT model, we have:
\begin{align*}
\resizebox{0.46\textwidth}{!}{
    $p^*(y_w \succ y_l|x)=\sigma\left(\beta\log\frac{\pi^*(y_w|x)}{\pi_\text{ref} (y_w|x)} - \beta\log\frac{\pi^*(y_l|x)}{\pi_\text{ref} (y_l|x)}\right)$.}
\end{align*}
We can then formulate a maximum likelihood estimation objective for the policy model $\pi_\theta$ on the preference dataset $\mathcal{D}$, resulting in the following training objective:
\begin{equation*}
    \mathcal{L}_\text{DPO}=-\mathbb{E}_{(x, y_w, y_l)\sim \mathcal{D}} \left[ \log p(y_w \succ y_l|x)\right].
\end{equation*}
% \sz{I guess either way works. which one you like the better}
% \begin{align*}
% \resizebox{0.46\textwidth}{!}{
%     $p^*(y_w \succ y_l \mid x)=\beta\log\frac{\pi^*(y_w\mid x)}{\pi_\text{ref} (y_w \mid x)} - \beta\log\frac{\pi^*(y_l\mid x)}{\pi_\text{ref} (y_l \mid x)}$.}
% \end{align*}
% \wz{Is this p a prob?}
% \begin{equation*}
%     \mathcal{L}_\text{DPO}=-\mathbb{E}_{(x, y_w, y_l)\sim \mathcal{D}} \left[ \log \sigma(p(y_w \succ y_l \mid x))\right].
% \end{equation*}
Here, the loss is calculated based on a uniform sampling of the preference dataset. In practice, the $y_w$ and $y_l$ in the preference dataset may be generated either with the same policy model being optimized, which corresponds to an on-policy RL setting~\cite{xu2023some, yuan2024self, rosset2024direct}, or with other models (e.g., GPT-4; \citealt{achiam2023gpt}), corresponding to an off-policy RL setting~\cite{tunstall2023zephyr, ivison2023camels, pal2024smaug}.

\subsection{Weighted Preference Optimization}
\label{ssec:wpo}

DPO does not require actively generating new outputs from the current policy, making it more cost-effective and suitable for off-policy settings.
However, DPO introduces a notable discrepancy between the distribution of outputs produced by the policy and those present in the preference dataset.
This divergence can lead to less effective learning.
% \sx{why do we need to introduce the notations here? they are used for helping the explanation.}
To illustrate, consider two instances of preference data: $\left(x^{(1)}, y_w^{(1)}, y_l^{(1)}\right)$ and $\left(x^{(2)},y_w^{(2)}, y_l^{(2)}\right)$, where the first tuple is sampled directly from the current policy model, while the second tuple is sampled from a different distribution from the current policy model. 
% \cz{of what? and the first tuple may also happen to be from a region of low probability}
Despite this difference in sampling probability, DPO treats both instances equally in its loss calculation, ignoring the fact that the first tuple, representing a more probable output of the current policy, should ideally exert a greater influence on the optimization process.
This oversight can lead to suboptimal performance, as DPO does not prioritize learning from the most representative or probable output of the policy model.
% Although some efforts conduct on-policy RL and generate outputs with the reference model~\cite{xu2023some,yuan2024self,rosset2024direct}, during the optimization process of policy model, the current policy and the reference model will diverge and introduce distribution shift.

% \cz{I suggest framing your approach as simulating on-policy RL using off-policy data, thus being both fast and enjoying the benefits from on-policy}

To address this issue, we propose to simulate on-policy RL using off-policy data, thereby being both fast and enjoying benefits from on-policy RL.

\smallskip
\noindent\textbf{Theoretical derivation.}
To simulate on-policy RL, we first transform the (off-policy) preference dataset $\mathcal{D}=\{(x^{(i)}, y_w^{(i)}, y_l^{(i)})\}_{i=1}^N$ into the following preference labeling function:
\begin{equation*}
    f(x, y_1, y_2)=\begin{cases}
        y_1 \succ y_2,& (x, y_1, y_2)\in \mathcal{D}\\
        y_2 \succ y_1,& (x, y_2, y_1)\in \mathcal{D}\\
        \textsc{na},&  \text{otherwise}\\
    \end{cases}
\end{equation*}
where we assume that the dataset contains no conflicting preferences, meaning that for any $x$, if $(x, y_1, y_2)\in \mathcal{D}$, then $(x, y_2, y_1)\notin \mathcal{D}$.
% \cz{what if both directions appear in D?}
We then conceptually generate a new preference dataset through a bootstrapping approach without actually carrying out the procedure.
Suppose an input $x$ is uniformly sampled from the original preference dataset, and then a pair of outputs $y_1, y_2$ is sampled with the current policy model. 
We retain the pair if it can be labeled by the labeling function, and otherwise reject the pair when $f(x,y_1,y_2)=\textsc{na}$.
% \cz{this is exactly what boostrapping is doing, so recommend using that terminology.}
%This process produces a new preference dataset with the same set of preference pairs but varying occurrences.
If we sample for an infinite amount of times, according to the law of large numbers, the occurrence rate of a pair $(x, y_w, y_l)$ would be proportional to $\pi_\theta(y_w|x)\pi_\theta(y_l|x) p(x)$.
% \cz{this prob needs to consider the empirical prob of x in existing dataset. Also need to say it's proportional to, as the integration is not 1.}
We then apply DPO to the newly generated preference dataset.

\smallskip
\noindent\textbf{Practical implementation.}
The conceptual process above is equivalent to optimizing the following weighted preference optimization (WPO) objective, where different pairs in the original preference dataset are reweighed:
% \cz{equation missing sigma, compared with DPO}
\begin{align}
\label{eq:wpo_loss}
\resizebox{0.43\textwidth}{!}{
    $\mathcal{L}_\text{WPO}=-\mathbb{E}_{(x, y_w, y_l)\sim \mathcal{D}} \left[ w(x,y_w) w(x,y_l) \log p(y_w \succ y_l|x)\right],$}
\end{align}
where $w(x,y)=\pi_\theta(y|x)$ and is \textit{detached} from back propagation.
Through this process, we effectively adjust the importance of each pair in the training process, giving greater weight to those pairs that are more likely to be sampled from the policy model, thus simulating on-policy RL.

In language models where $y_w$ and $y_l$ are sequences of tokens, the product of the conditional probabilities $\pi_\theta(y_w|x) \cdot \pi_\theta(y_l|x)$ can be exceedingly small and exhibit high variance among different pairs.
To address this, we utilize the length-normalized sequence probability as a weighting factor:
\begin{equation*}
w(x, y) = \exp\left(\frac{1}{|y|} \sum_{t=1}^{|y|} \log \pi_\theta(y_t|x, y_{<t})\right),
\end{equation*}
where $|y|$ represents the number of tokens in the output.

\subsection{Weight Alignment}
\begin{figure}
    \centering
    \includegraphics[width=0.88\linewidth]{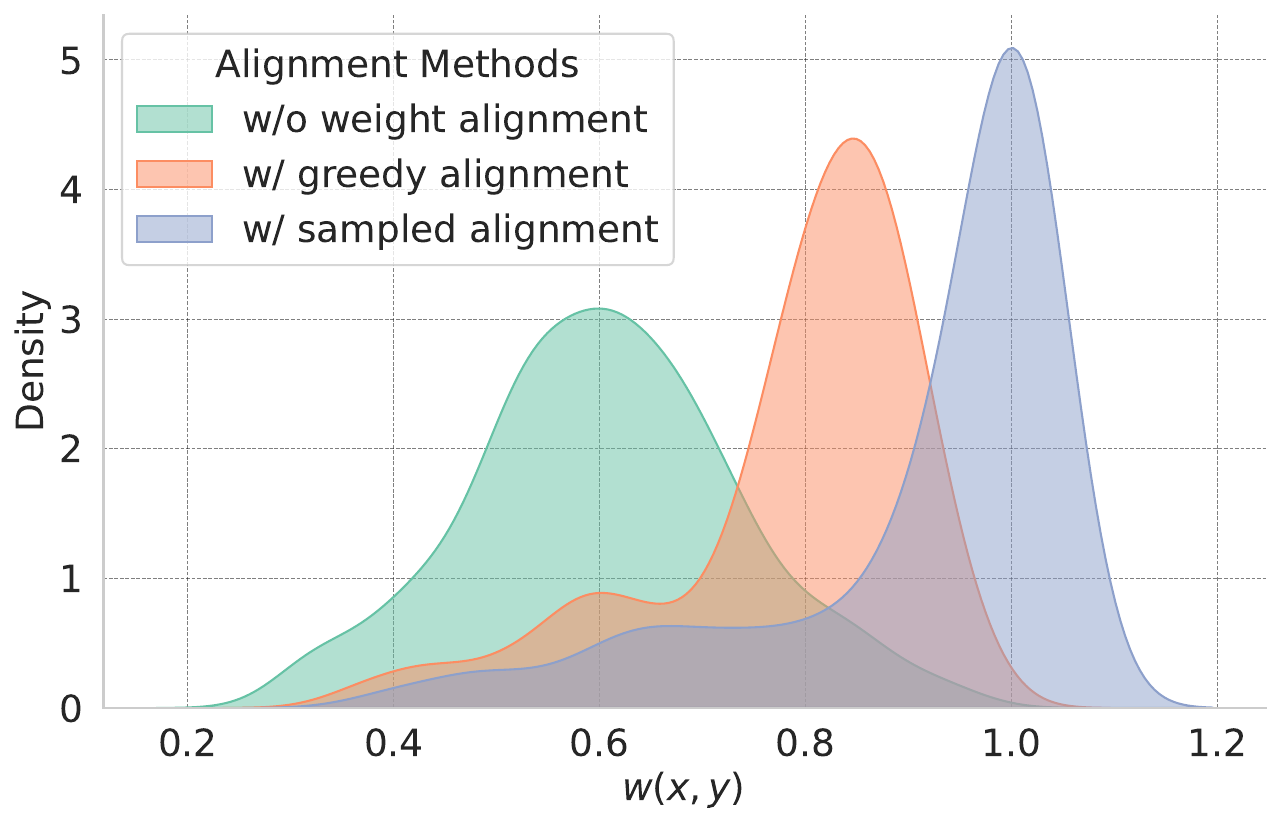}
    \caption{Weight distribution of outputs sampled using the policy model with different alignment methods.}
    \label{fig:weight_alignment}
    \vspace{-15pt}
\end{figure}
\label{ssec:weight_alignment}
The objective of our weighting strategy is to simulate on-policy RL, where outputs are weighted according to how closely they align with on-policy behavior.
For outputs generated by the current policy model, we expect their weights to be uniformly 1, while outputs that deviate from this on-policy behavior should receive smaller weights.
However, due to the varying levels of confidence that LLMs exhibit across different inputs~\cite{si2023prompting,xiong2024can}, even outputs generated by the policy model may sometimes be assigned low weights.
This introduces an unintended bias where some on-policy outputs receive lower weights purely because of lower model confidence based on the input, disrupting the uniformity we aim to achieve.
Figure~\ref{fig:weight_alignment} shows the weight distribution of sampled outputs based on prompts from Ultrafeedback and the Mistral-sft-beta model, in which we observe significant variability in $w(x,y)$.
To address this and ensure equal weighting of these outputs, we propose to align the weights in WPO.

A direct method is to adjust the weights above by the sequence probability of the on-policy outputs sampled from the policy model.
% \cz{what is weight of onpolicy pair?}
However, generating outputs during training is computationally expensive, and hence, we explore approximation methods for this alignment.
Instead of using weights of the whole sequences as reference, we operate at the token level and adjust the probability of output tokens according to the token distribution in the policy model, based on the current subsequence. We propose two ways to achieve the alignment.

\smallskip
\noindent \textbf{Greedy alignment.}
In this approach, we adjust the weights based on greedy decoding by comparing the probability of the current token with that of the most probable token in the vocabulary.
Specifically, we adjust weights based on the maximum token probability among the set of all tokens in the subsequence, defined as:
\begin{equation*}
\resizebox{0.46\textwidth}{!}{
    $w(x, y)=\exp\left(\frac{1}{|y|}\sum\limits_{t=1}^{|y|}\log \frac{\pi_\theta(y_t|x, y_{<t})}{\max_{v\in \mathcal{V}}\pi_\theta(v|x, y_{<t})}\right),$}
\end{equation*}
where $\mathcal{V}$ represents the set of all tokens in the language model.

\smallskip
\noindent \textbf{Sampled alignment.}
In this approach, we adjust weights based on outputs that are randomly sampled from the policy model at a temperature of 1.0.
Since the probability for each token $v$ is computed as $\pi_\theta(v|x, y_{<t})$, the expected probability of a randomly sampled token would be $\sum_{v\in \mathcal{V}}\pi_\theta(v|x, y_{<t})^2$, and the calibrated weights are then given by:
\begin{align}
\label{eq:weight_sample}
\resizebox{0.40\textwidth}{!}{
    $w(x, y)=\exp\left(\frac{1}{|y|}\sum\limits_{t=1}^{|y|}\log \frac{\pi_\theta(y_t|x, y_{<t})}{\sum_{v\in \mathcal{V}}\pi_\theta(v|x, y_{<t})^2}\right).$}
\end{align}
We use sampled alignment as the default alignment method in WPO due to its superior performance, as confirmed in Section~\ref{ssec:main_results}.
Additionally, in Figure~\ref{fig:weight_alignment}, sampled alignment leads to a more concentrated weight distribution of outputs from the policy model, thereby better simulating on-policy RL.

\section{Experiment}
In this section, we outline our experimental settings (Section~\ref{ssec:experiment_settings}) and present the main results along with ablation studies (Section~\ref{ssec:main_results}).
We then compare different RL settings (Section~\ref{ssec:rl_settings}). Additional analysis of WPO is provided in Appendix~\ref{ssec:analysis}.

\begin{table*}[!t]
    \centering
    {
    \scalebox{0.77}{
    \begin{tabular}{llcccccccc}
    \toprule
     \multicolumn{2}{c}{\multirow{4}{*}{\textbf{Method}}}& \multicolumn{4}{c}{\textbf{Mistral-Base (7B)}}& \multicolumn{4}{c}{\textbf{Llama-3-Instruct (8B)}}\\
     \cmidrule(lr){3-6}\cmidrule{7-10}
     &&\multicolumn{2}{c}{\textbf{Alpaca Eval 2.0}}& \multicolumn{2}{c}{\textbf{MT-bench}}& \multicolumn{2}{c}{\textbf{Alpaca Eval 2.0}}& \multicolumn{2}{c}{\textbf{MT-bench}} \\
     \cmidrule(lr){3-4}\cmidrule(lr){5-6}\cmidrule(lr){7-8}\cmidrule{9-10}
     &&Len-control.& Win Rate& Avg.&Win Rate& Len-control.& Win Rate& Avg.&Win Rate\\
     &&Win Rate& vs GPT-4& Score& vs DPO& Win Rate& vs GPT-4& Score& vs DPO \\
     \midrule
     &SFT& 9.5& 5.8& 6.64&-& 26.0& 25.3& 7.97& - \\
     \midrule
     \parbox[t]{2mm}{\multirow{5}{*}{\rotatebox[origin=c]{90}{Off-policy}}}& ORPO& 14.7& 12.6& 7.32&-& -& -&-& -\\
     % \parbox[t]{2mm}{\multirow{5}{*}{\rotatebox[origin=c]{90}{Off-policy}}}& ORPO& 14.7& 12.6& 7.32&-& 20.6& 18.2&7.79& -\\
     &KTO& 14.9& 12.3& 7.36&-& -& -&-&-\\
     &SimPO& 21.5& 21.4& 7.32& -& -& -& -& -\\
     &DPO& 20.6 (0.7)& 18.6 (1.0)& 7.36 (0.04)&50 (0)& 28.2 (0.5)& 24.0 (0.5)& 8.10 (0.05)& 50 (0)\\
     % &WPO \textit{w/o align.}& \underline{22.02 (1.09)}& 20.29 (1.77)& 1802& 7.16 (0.21)& 54.4 (4.9) \\
     % &WPO \textit{w/ greedy align.}& \underline{23.04 (1.29)}& \underline{21.39 (1.93)}& 1825& 7.34 (0.04)&\underline{57.9 (5.6)} \\
     &WPO& \underline{\textbf{24.4 (1.4)}}& \underline{\textbf{23.7 (2.1)}}& \textbf{7.37 (0.10)}&\underline{\textbf{60.1 (4.7)}}& \underline{\textbf{33.8 (1.3)}}& \underline{\textbf{31.0 (1.8)}}& \textbf{8.14 (0.05)}& \underline{\textbf{58.1 (3.4)}} \\
     \midrule
     \parbox[t]{2mm}{\multirow{3}{*}{\rotatebox[origin=c]{90}{Hybrid}}}&DPO & 37.9 (1.2) & 40.3 (1.1)& 7.14 (0.41)& 50 (0)& 44.2 (1.2)& 48.6 (1.0)& 8.16 (0.10)& 50 (0) \\
     % &\textit{+ Ultrafeedback}& \underline{40.84 (1.02)}& \underline{45.03 (2.87)}& 2696& 7.27 (0.03)& 51.1 (5.3) \\
     &WPO & \underline{42.0 (1.7)}& \underline{46.2 (2.3)}& \textbf{7.38 (0.08)}& \underline{56.4 (4.6)}& 45.8 (1.3)& 50.0 (1.1)& \textbf{8.18 (0.22)}& \underline{54.8 (2.2)} \\
     &\textit{+ Ultrafeedback}& \underline{\textbf{43.1 (1.1)}}& \underline{\textbf{49.6 (1.2)}}& 7.23 (0.19)& \underline{\textbf{58.8 (4.5)}}& \underline{\textbf{48.6 (1.3)}}& \underline{\textbf{52.1 (1.2)}}& 8.14 (0.10)& \underline{\textbf{55.1 (2.4)}} \\
     \bottomrule
        \end{tabular}}}
    \caption{Alpaca Eval 2.0 and MT-bench results. We report the average and standard deviation of the results from 5 runs of different random seeds. Scores that are \underline{underlined} denote statistically significant gains ($p < 0.05$).
    % \cz{add some other baseline models better than DPO. Second turn result not good, delete?}
    }
    \label{tab:main_result}
    \vspace{-15pt}
\end{table*}

\subsection{Experimental Settings}
\label{ssec:experiment_settings}
\smallskip
\noindent\textbf{Model configurations.} Our methods are implemented based on the official code of zephyr\footnote{\url{https://github.com/huggingface/alignment-handbook}}.
For Mistral-base, we adopt the official hyperparameters from zephyr.
Specifically, we use the SFT checkpoint of zephyr\footnote{\url{https://huggingface.co/HuggingFaceH4/mistral-7b-sft-beta}} as our SFT model.
Training is conducted over a single epoch with a batch size of 128, a learning rate of $5\text{e-}7$, a warm-up phase for 10\% of the training, and a cosine decay schedule. We set $\beta$ to 0.01 for both DPO and WPO.
For Llama-3-Instruct, we perform a hyperparameter search within the range recommended by~\citet{meng2024simpo}. Our final hyperparameters are a learning rate of $1\text{e-}6$, two training epochs, and $\beta$ of 0.01 for both DPO and WPO. For all training configurations, we conduct training for 5 runs with different random seeds and report both the average results and their standard deviation.

\smallskip
\noindent\textbf{Training data.}
We perform RLHF in off-policy and hybrid settings.
% \cz{need to mention in method section that WPO can also be applied to on-policy, and what does that mean probabilistically? (won't it have square of pi(y|x), sampled + weight?)}
In the \textit{off-policy} setting, we use the binarized Ultrafeedback dataset\footnote{\url{https://huggingface.co/datasets/HuggingFaceH4/ultrafeedback_binarized}}\cite{cui2023ultrafeedback}, which compromises 63k preference pairs sampled from models other than our SFT model, such as GPT-4 and Llama-2~\cite{touvron2023llama}.
In the \textit{hybrid} setting, we follow the approach in DNO~\cite{rosset2024direct}, using data generated from both the policy model and other models.
Specifically, we sample 5 outputs from the SFT model based on prompts from Ultrafeedback and add another output generated by \texttt{gpt-4-turbo}.
We employ top-p sampling with $p=0.95$ and a temperature of 0.7. Preference annotations are produced using \texttt{gpt-4-turbo} with additive scoring prompt.
For each prompt, we select outputs scoring 5 or 6 as $y_w$ and then choose a random output with a score at least one point lower as $y_l$.
If such a pair cannot be found, the prompt is not used.
This data construction step produces a smaller preference dataset, so we further employ the \textit{+ Ultrafeedback} setting, where we add the missing prompts back using the preference pairs from Ultrafeedback.

\smallskip
\noindent \textbf{Evaluation.}
We evaluate the models on Alpaca Eval 2 and MT-bench.
Alpaca Eval 2 is an automated metric that measures LLMs' alignment with human preferences using 805 representative instructions. For each instruction, the evaluated model's response and \texttt{gpt-4-turbo}'s response are compared head-to-head using an auto-evaluator. The win rate is the probability that the auto-evaluator prefers the evaluated model's responses. Alpaca Eval 2 also introduces a length-controlled win rate~\cite{dubois2024length} to address the length bias of \texttt{gpt-4-turbo}. We follow the generation configurations in~\citet{tunstall2023zephyr} for Mistral models and in~\citet{zheng2024weak} for Llama-3 models.

MT-bench is an LLM-based automated evaluation metric comprising 80 challenging questions.
We report results using two scoring methods. In the single answer grading approach, the auto-evaluator (\texttt{gpt-4-0613}) assigns scores from 1 to 10 to responses, and we report the average scores. In the pairwise comparison approach, the evaluator (\texttt{gpt-4-0613}) compares two responses to decide which is better or if it's a tie (recorded as 0.5 in win rate). The pairwise method can detect more subtle differences between responses than single answer grading. We use the official generation configurations in MT-bench.

% \begin{itemize}
%     \item On-policy RL (Iteration 1): sample outputs from SFT model (HuggingFaceH4/mistral-7b-sft-beta), construct preference pairs by GPT-4.
%     \item Off-policy + On-policy RL (Iteration 1): finetune the model with both off-policy and on-policy data. Need a hyperparameter to tune the weight of these two parts.
% \end{itemize}

\subsection{Main Results and Ablation}
\label{ssec:main_results}

\smallskip
\textbf{WPO consistently and significantly outperforms DPO and its variants.}
The main results are shown in Table~\ref{tab:main_result}.
We include the results of different preference optimization algorithms such as DPO, ORPO~\cite{hong2024orpo}, KTO~\cite{ethayarajh2024kto}, and SimPO~\cite{meng2024simpo} on the two benchmarks.
For ORPO, KTO, and SimPO, we report the evaluation results of their official model checkpoints on Mistral-base.\footnote{We do not include their results on Llama-3-Instruct in the off-policy setting as the official checkpoints are unavailable. Reproducing these methods requires extensive hyperparameter searches, which may not yield the optimal hyperparameter values for a fair comparison.}
We find that WPO generally outperforms DPO in all settings and also outperforms all its variants on Mistral-base in the off-policy setting.
Particularly, when trained with the Llama-3-Instruct model and the hybrid \textit{+Ultrafeedback} setting, WPO achieves a new state-of-the-art length-controlled win rate of 48.6\% against GPT-4-turbo on Alpaca Eval 2.
These results highlight the effectiveness of WPO.
Additionally, while DPO underperforms compared to SimPO, it still demonstrates competitive results, providing a solid basis for WPO.

\smallskip
\noindent\textbf{Varied separation of benchmarks.}
On MT-bench, the average score does not effectively distinguish the performance of different models.
Additionally, we observe variability in the average MT-bench score. Even when using GPT-4 to score the same outputs with a temperature of 0, the score can vary by up to 0.1 at different times.
Given the clearer separation in our experiments and the greater alignment with human evaluations, as shown in the original paper~\cite{zheng2024judging}, we consider pairwise win rate to be a more suitable metric for assessing different alignment methods.
Therefore, we use it for MT-bench in the following part of the paper.

\begin{table}[]
    \centering
    \scalebox{0.72}{
    \begin{tabular}{lccc}
    \toprule
        \multirow{3}{*}{\textbf{Method}}& \multicolumn{2}{c}{\textbf{Alpaca Eval 2.0}} & \textbf{MT-bench}\\
        \cmidrule(lr){2-3}\cmidrule(lr){4-4}
        &Len-control.& Win Rate& Win Rate\\
        &Win Rate& vs GPT-4& vs DPO \\
        \midrule
        WPO \textit{w/ sampled align.}& 24.4& 23.7& 60.1\\
        \midrule
        WPO \textit{w/ greedy align.}& 23.0& 21.4&57.9 \\
        WPO \textit{w/o align.}&22.0& 20.3& 54.4 \\
        DPO& 20.6 & 18.6& 50 \\
        \bottomrule
    \end{tabular}}
    \caption{Ablation of weight alignment methods on Mistral-base in the off-policy setting. sampled alignment, the default weight alignment method, yields the best results.}
    \label{tab:ablation}
    \vspace{-10pt}
\end{table}

\smallskip
\noindent\textbf{Sampled weight alignment works the best.}
Table~\ref{tab:ablation} shows the results of WPO with different weight alignment methods on Mistral-base in the off-policy setting.
We observe that sampled alignment outperforms other variations on both benchmarks, while greedy sampling outperforms w/o alignment.
We also find that the ranking of performance matches the ranking of concentration levels in the weight distribution shown in Figure~\ref{fig:weight_alignment}.
This indicates that weight alignment enables a more effective simulation of on-policy RL, leading to improved performance.

\begin{table}[]
    \centering
    \scalebox{0.77}{
    \begin{tabular}{lccc}
    \toprule
        \multirow{3}{*}{\textbf{Method}}& \multicolumn{2}{c}{\textbf{Alpaca Eval 2.0}} & \textbf{MT-bench}\\
        \cmidrule(lr){2-3}\cmidrule(lr){4-4}
        &Len-control.& Win Rate& Win Rate\\
        &Win Rate& vs GPT-4& vs Baseline \\
        \midrule
        IPO & 25.0& 21.2& 50 \\
        SimPO& 21.5& 21.4& 50 \\
        KTO & 14.9& 12.3& 50 \\
        \midrule
        WPO$_\textsc{ipo}$& 29.4& 25.7& 54.1  \\
        WPO$_\textsc{simpo}$& 21.9& 24.6& 52.5\\
        WPO$_\textsc{kto}$& 21.1& 20.3& 60.0 \\
        \bottomrule
    \end{tabular}}
    \caption{Results of WPO with different loss functions for preference optimization on Mistral-base in the off-policy setting, which show that incorporating WPO leads to consistent improvements.}
    \label{tab:analysis_losses}
    \vspace{-10pt}
\end{table}

\smallskip
\noindent\textbf{WPO also improves other loss functions for preference optimization.}
It is important to note that, in addition to DPO, there are other loss functions for aligning LLMs.
Since WPO works by weighing preference data and is independent to the loss function being used, it can be easily integrated into them.
We investigate whether the integration of WPO enhances the performance of other loss functions.
Existing losses can be categorized into those using paired preference data and those utilizing unpaired preference data.
For losses using paired data, we weigh each pair similarly to DPO. For losses using unpaired data, we weigh each output $y$ independently with $w(x, y)$ and normalize the weights so that the total weights of favored outputs and disfavored outputs are both 1 within the batch. This normalization ensures a balance between favored and disfavored outputs in the loss.
In this study, we considered IPO~\cite{azar2024general} and SimPO for alignment with paired data, and KTO for alignment with unpaired data.
The results on Mistral-base in the off-policy setting, shown in Table~\ref{tab:analysis_losses}, indicate that integrating WPO leads to improved results for all loss functions.
This demonstrates that WPO provides universal improvements across different loss functions for preference optimization.

\smallskip
\noindent\textbf{Better base and reward models yield stronger results.}
To further enhance our model, we investigate using better base models and reward models.
Specifically, we adopt the Gemma-2-9b-it~\cite{team2024gemma} as the base model.
In a setup similar to our hybrid approach, we sample five outputs from Gemma and one additional output from \texttt{gpt-4-turbo}.
To rank these outputs, we apply ArmoRM~\cite{wang2024arithmetic,ArmoRM} and use the best and worst outputs to form preference pairs.
We use the same set of training hyperparameters used in Llama-3-Instruct.
The model finetuned using WPO achieves a length-controlled win rate of 76.7\% and a win rate of 77.8\% on Alpaca Eval 2, demonstrating the effectiveness of this approach.

\subsection{Comparison of Different RL Settings}
\label{ssec:rl_settings}
\begin{figure}
    \centering
    \includegraphics[width=0.95\linewidth]{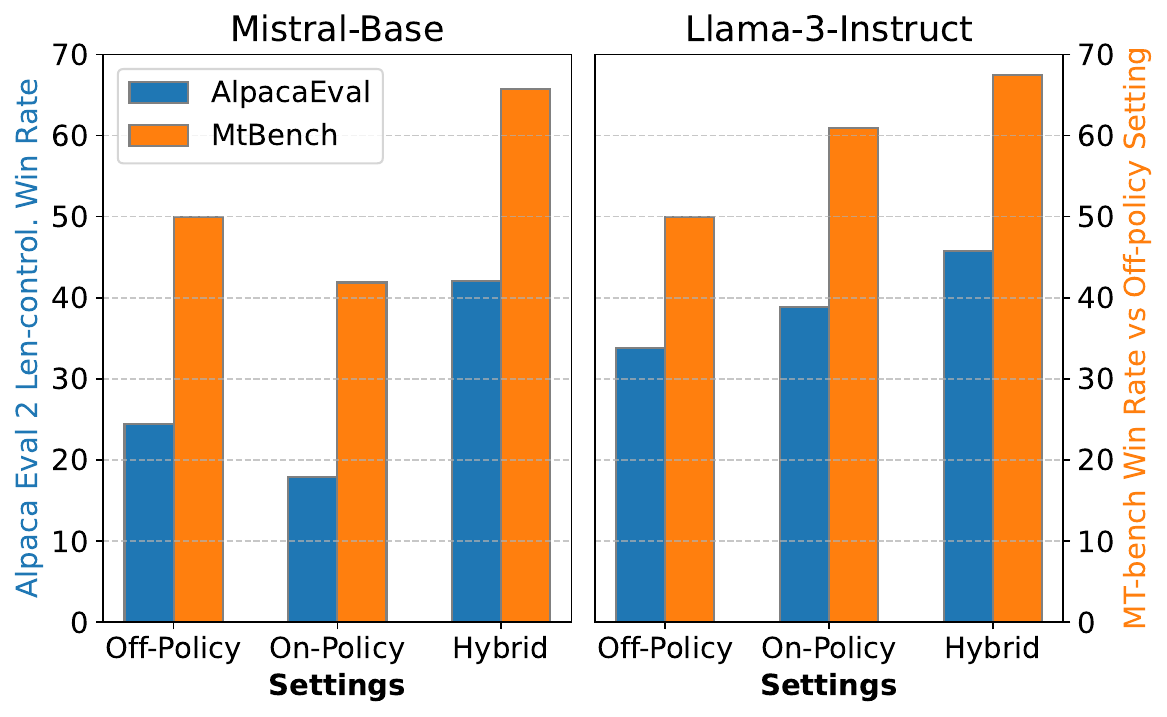}
    \caption{Results of WPO in different RL settings. The hybrid setting consistently yileds better results than other RL settings.}
    \label{fig:rl_comparisions}
    \vspace{-10pt}
\end{figure}

Recent studies on RLHF have employed various RL settings where preference data is generated in an off-policy, on-policy, or hybrid manner.
Existing work~\cite{tang2024understanding,xu2024dpo} has demonstrated that on-policy preference optimization outperforms off-policy methods, while~\citet{rosset2024direct} show that incorporating high-quality off-policy outputs can yield superior performance, as these outputs can introduce valuable information that the current policy might not encounter on its own.
In this study, we compare model performance trained with WPO across these RL settings. The results are presented in Figure~\ref{fig:rl_comparisions}, showcasing the length-controlled win rate on Alpaca Eval 2 and the pairwise win rate compared to the off-policy setting on MT-bench.

\smallskip
\noindent\textbf{Hybrid RL achieves the best results.}
Figure~\ref{fig:rl_comparisions}
% \sz{do we have the results or references of this in the paper. For example "confirms our results in Section 5.1.1" etc}
shows that for both Mistral-base and Llama-3-Instruct, the hybrid setting—utilizing both on-policy data and high-quality off-policy data from \texttt{gpt-4-turbo}—consistently delivers superior performance.
This suggests that combining high-quality off-policy data and on-policy data can significantly enhance preference optimization, which is consistent to the results in~\citet{rosset2024direct}.

\smallskip
\noindent\textbf{On-policy is not always better than off-policy.}
Our analysis reveals that the effectiveness of on-policy versus off-policy preference optimization is model-dependent \cite{munos2016safe, voloshin2019empirical}.
For the Mistral-base model, off-policy setting yields slightly better performance, while for Llama-3-Instruct, on-policy setting shows better performance.
We attribute this variation to the quality of the SFT model.
In the case of Mistral-base, the sampled outputs are of lower quality, causing the preference optimization process to mimic suboptimal outputs and leading to poorer results.
This highlights the importance of the initial policy's quality and suggests that models with higher initial performance might benefit more from on-policy optimization, while those with lower initial quality may not gain as much.

\smallskip
\noindent\textbf{The dispreferred data should be on-policy, the preferred data benefits less.}
While WPO simulates on-policy data by weighing both $y_w$ and $y_l$ in the preference data, these two outputs play different roles during optimization.
The gradient of the WPO is given by:
\begin{align*}
&\resizebox{0.9\hsize}{!}{$\nabla \mathcal{L}_{\mathrm{WPO}}=-\beta w(x,y_w) w(x,y_l) \sigma\left(\hat{r}\left(x, y_l\right)-\hat{r}\left(x, y_w\right)\right)$} \\
&\resizebox{0.9\hsize}{!}{$[\underbrace{\nabla \log \pi\left(y_w|x\right)}_{\text {increase the probability of } y_w}-\underbrace{\nabla \log \pi\left(y_l|x\right)}_{\text {reduce the probability of } y_l}]$}.
\end{align*}
That is, WPO will make the policy model mimic $y_w$ while moving away from $y_l$.
Given their different optimization directions, we investigate the importance of on-policy sampling for $y_w$ and $y_l$ in preference optimization.
To achieve this, we further study two different variants of WPO, namely WPO$_\textsc{w}$ and WPO$_\textsc{l}$.
These losses are formulated as follows:
\begin{minipage}{\linewidth}
\centering
\resizebox{0.92\textwidth}{!}{
\begin{minipage}{\textwidth}
\begin{align*}
    \mathcal{L}_{\text{WPO}}&= -\mathbb{E}_{(x, y_w, y_l)\sim \mathcal{D}} \left[ w(x,y_w) w(x,y_l) \log p(y_w \succ y_l|x)\right],\\
    \mathcal{L}_{\text{WPO}_\textsc{w}}&= -\mathbb{E}_{(x, y_w, y_l)\sim \mathcal{D}} \left[ w(x,y_w) \log p(y_w \succ y_l|x)\right],\\
    \mathcal{L}_{\text{WPO}_\textsc{l}} &= -\mathbb{E}_{(x, y_w, y_l)\sim \mathcal{D}} \left[ w(x,y_l) \log p(y_w \succ y_l|x)\right],\\
\end{align*}
\end{minipage}
}
\end{minipage}

\begin{figure}
    \centering
    \includegraphics[width=0.95\linewidth]{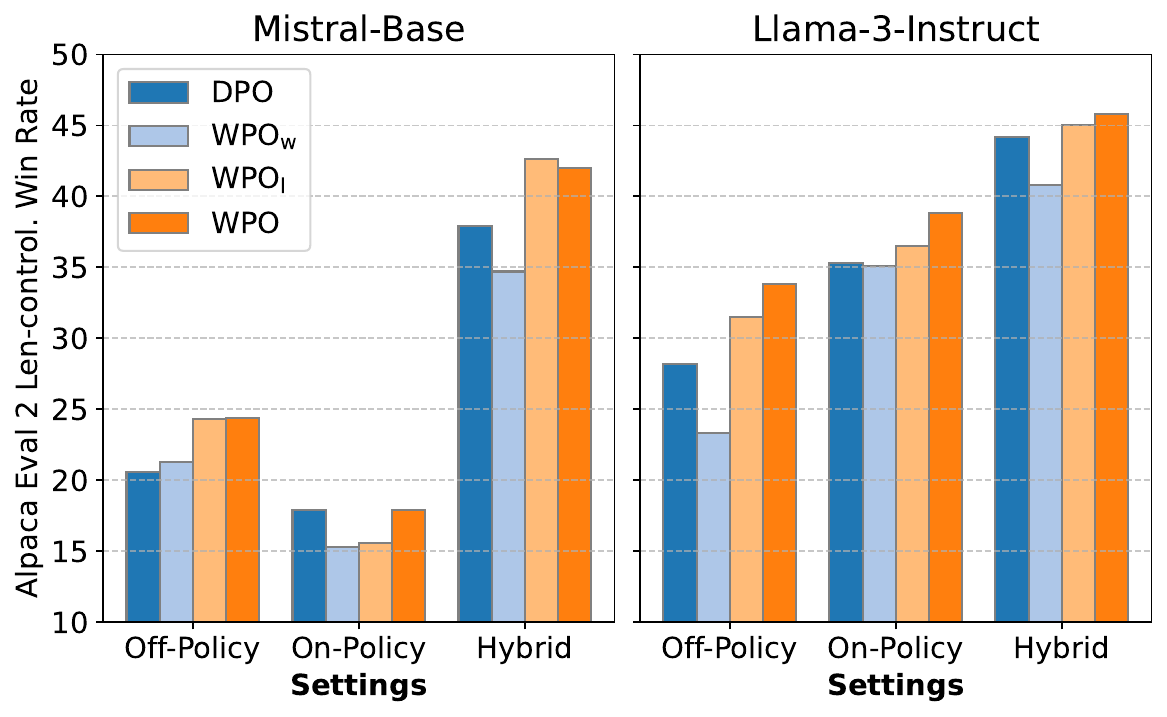}
    \caption{Results of variations of WPO in different RL settings.}
    \label{fig:wpo_variations}
    \vspace{-10pt}
\end{figure}

where in WPO$_\textsc{w}$, we only increase the weights of pairs where $y_w$ is more closed to on-policy outputs. For WPO$_\textsc{l}$, we only increase the weights of pairs where $y_l$ is closer to on-policy outputs.
% \cz{this is confusing from math perspective, as it completely flips the sign of log pi(yl|x) from the above equation, and the performance is similar..}
Results on Mistral-base and Llama-3-Instruct are in Figure~\ref{fig:wpo_variations}. It shows that WPO$_\textsc{l}$ generally achieves similar results to WPO.
Conversely, WPO$_\textsc{w}$ consistently underperforms WPO and even underperforms DPO in most settings.
Therefore, making $y_l$ on-policy explains most of the improvements of WPO, while making $y_w$ on-policy is still useful but not as important.
This finding suggests that using on-policy, dispreferred data is important for preference optimization, while using on-policy preferred data may be beneficial but not as critical.
% \sz{I think this paragraph is also confusing. My understanding is that for preferred data does need on-policy weight. Instead of only having the above DPO gradient, we can add original WPO loss instead of only having  $\mathcal{L}_{\text{WPO}_\textsc{w}}$}

\section{Conclusion}
In this study, we tackled the distributional gap problem inherent in off-policy preference optimization.
By introducing Weighted Preference Optimization (WPO), we successfully simulated on-policy preference optimization using off-policy preference data, merging the benefits of both approaches.
Our method not only addressed the distributional gap without incurring additional costs but also enhanced the effectiveness of preference optimization.
Extensive experiments demonstrate that WPO can produce better LLMs that are more closely aligned with human preferences.

\section*{Limitations}
\smallskip
\noindent \textbf{The performance gap between off and on-policy preference optimization remains.} Although WPO simulates on-policy RL with off-policy data, it does not fully bridge the performance gap between off-policy and on-policy RL. As shown in the results, even with WPO, off-policy methods may still underperform compared to on-policy and hybrid methods. Therefore, while we propose WPO as a solution, it does not entirely eliminate the performance disparity, and on-policy preference data remains important.
Future work will be on how to further reduce this performance gap without incurring additional training costs.

\smallskip
\noindent \textbf{Comprehensiveness of preference dataset.} The goal of our experiments is to compare WPO with other preference optimization algorithms, not to provide a comprehensively aligned LLM. In our experiments, we use Ultrafeedback as the preference data, which primarily focuses on helpfulness, truthfulness, and instruction following, and does not include safety aspects. Additionally, it does not consider preference optimization for multi-turn conversations. Future work should involve collecting more comprehensive preference datasets and integrating multiple aspects of preference optimization to train better-aligned LLMs.

\bibliography{custom}

\onecolumn
\appendix

\begin{table*}[!t]
\centering
\scalebox{0.88}{
    \begin{tabular}{lccccccc}
    \toprule
    Method& ARC& TruthfulQA& WinoGrande& GSM8k& HellaSwag& MMLU& Average \\
    \midrule
    \multicolumn{8}{c}{\textbf{Mistral-Base (7B)}} \\
    \midrule
    SFT& 58.19& 43.03& 77.51& 38.89& 82.30& 59.78& 59.95\\
    Off-policy DPO& 64.42& 52.44& 79.48& 30.17& 85.36& 59.78& 61.94\\
    Off-policy WPO& 64.08& 51.07& 78.14& 32.60& 85.17& 59.51& 61.76 \\
    Hybrid DPO& 64.76 & 60.46 & 78.22 & 32.15 & 85.30 & 58.75 & 63.27  \\
    Hybrid WPO& 65.70 & 57.62 & 79.08 & 30.71 & 85.15 & 59.82 & 63.01 \\
    \midrule
    \multicolumn{8}{c}{\textbf{Llama-3-Instruct (8B)}} \\
    \midrule
    SFT& 61.60 & 51.65 & 76.72 & 75.82 & 78.68 & 65.65 & 68.35\\
    Off-policy DPO& 68.00 & 61.07 & 77.43 & 74.68 & 82.26 & 66.31 & 71.63\\
    Off-policy WPO& 66.98 & 58.91 & 75.45 & 71.95 & 81.87 & 65.97 & 70.19\\
    Hybrid DPO& 65.53 & 56.10 & 78.93 & 75.13 & 81.12 & 65.72 & 70.42\\
    Hybrid WPO& 65.27 & 55.47 & 79.72 & 66.72 & 81.02 & 65.97 & 69.03 \\
    \bottomrule
    \end{tabular}}
    \caption{Results on the OpenLLM leaderboard.}
    \label{tab:general_results}
\end{table*}

\begin{figure}[!t]
    \centering
    \includegraphics[width=0.4\linewidth]{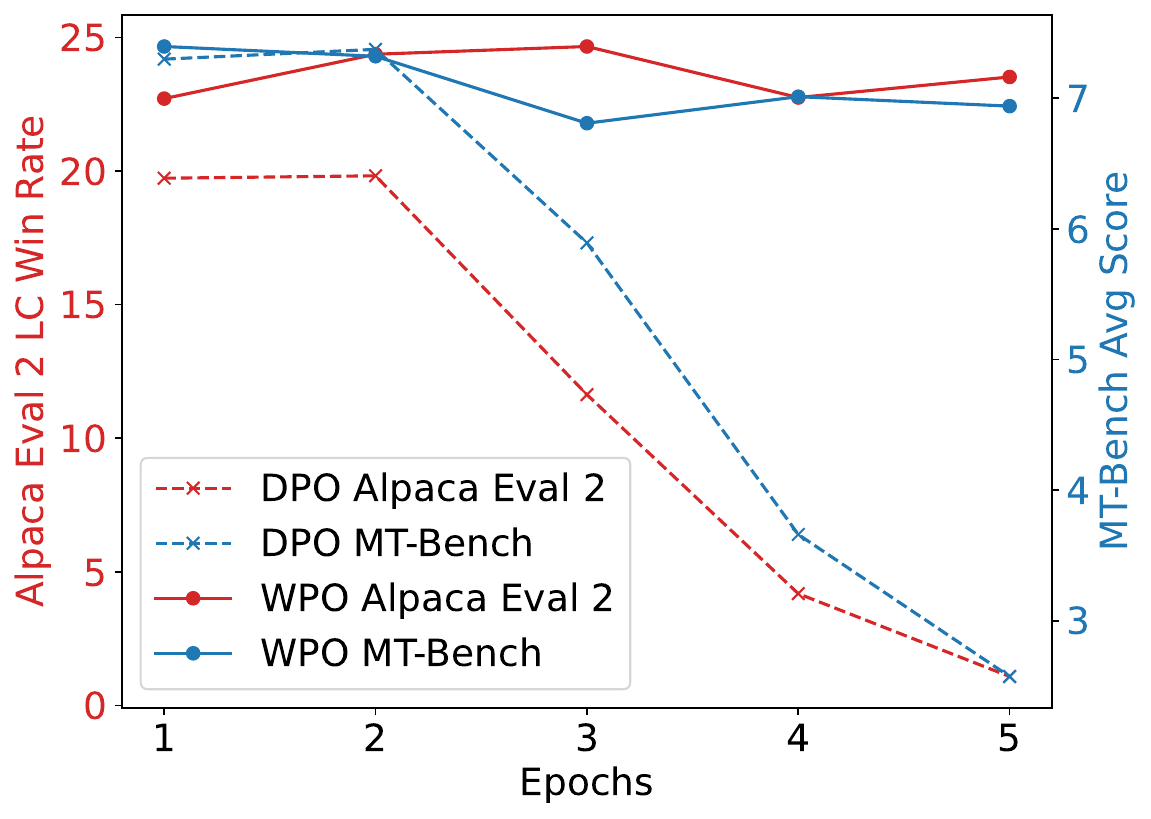}
    \caption{Results of DPO and WPO when trained for more epochs.}
    \label{fig:more_epochs}
\end{figure}

\section{Additional Analysis}
\label{ssec:analysis}
\smallskip
\noindent \textbf{Results on downstream tasks.} We further evaluate the performance of SFT, DPO, and WPO models on the OpenLLM leaderboard~\cite{open-llm-leaderboard} to assess their capabilities on downstream tasks. For this evaluation, we use the lm-evaluation-harness\footnote{\url{https://github.com/EleutherAI/lm-evaluation-harness}}, the official code base for the OpenLLM leaderboard. Results are shown in Table~\ref{tab:general_results}.
Generally, we find that preference optimization with DPO or WPO outperforms the SFT model, while Llama-3-Instruct based models outperform Mistral-base. However, we do not observe a correlation between performance on the OpenLLM leaderboard and performance on instruction-following benchmarks such as Alpaca Eval 2 and MT-bench. For example, although Llama-3-Instruct with DPO or WPO in the hybrid setting shows the best results on instruction-following benchmarks, it underperforms its off-policy counterparts on the OpenLLM leaderboard.
Additionally, we find that preference optimization may not improve results on all downstream tasks. On MMLU, the results are similar to SFT, and on GSM8K, the results are even lower than SFT in all settings. Our findings are consistent with the alignment tax phenomenon~\cite{askell2021general}, which indicates that better alignment may not improve and can sometimes even hurt performance on downstream tasks.

\smallskip
\noindent\textbf{Comparison between DPO and WPO on training dynamics.}
We investigate how the performance of DPO and WPO changes with different numbers of training epochs.
Both DPO and WPO were trained using the SFT checkpoint of Mistral-base and the Ultrafeedback dataset for five epochs, with evaluation results recorded at the end of each epoch, as shown in Figure~\ref{fig:more_epochs}.
In this study, we use the same set of hyperparameters as mentioned in Section~\ref{ssec:experiment_settings}, with DPO and WPO using the same set of hyperparameters.
We observed that DPO's performance declines sharply after two epochs, suggesting strong reward model overoptimization~\cite{rafailov2024scaling}. In contrast, WPO maintains consistent performance over more epochs, indicating better training stability. This suggests that simulating on-policy RL, as done by WPO, may mitigate issues related to reward model overoptimization and increase the stability of preference optimization.
Furthermore, a comparison of results between DPO and WPO, particularly on Alpaca Eval 2, shows that the peak performance of DPO across various epochs still falls below that of WPO. This indicates that WPO not only provides more stable training dynamics but also finds a different and better solution than DPO. This enhanced performance and stability highlight the advantages of WPO in effectively leveraging the preference data and maintaining stable and robust preference optimization throughout the training process.

\section{Link of Open Sourced Models in Experiments}
The list of open-sourced LLMs and their Huggingface IDs are listed in Table~\ref{tab:list_of_models}.
\begin{table}[!h]
    \centering
    \scalebox{0.88}{
    \begin{tabular}{ll}
    \toprule
    Model&  Huggingface ID\\
    \midrule
    Mistral-base SFT& \url{HuggingFaceH4/mistral-7b-sft-beta}\\
    Mistral-base ORPO& \url{kaist-ai/mistral-orpo-beta} \\
    Mistral-base KTO& \url{ContextualAI/zephyr_sft_kto} \\
    Mistral-base SimPO& \url{princeton-nlp/Mistral-7B-Base-SFT-SimPO} \\
    Llama-3-instruct SFT& \url{meta-llama/Meta-Llama-3-8B-Instruct} \\
    \bottomrule
    \end{tabular}}
    \caption{List of open-source models in experiments.}
    \label{tab:list_of_models}
\end{table}

\section{Additional Details}
\smallskip
\noindent\textbf{Scientific artifacts.}
We use various scientific artifacts throughout the paper, including base LLM models, preference datasets, and evaluation tools/benchmarks. References to all used artifacts are provided, and details such as their license, language, coverage, number of parameters, and any safety issues can be found by following the respective references. Note that current LLMs and preference datasets may encompass a wide range of data types and utilizes data from different domains and sources, so we do not list the details in this paper and encourage readers to refer to the original sources for more information. In this paper, we primarily use these artifacts for non-distributive and non-commercial purposes, which is in compliance with their licenses.

\smallskip
\noindent\textbf{Budget.}
We conduct all experiments using 8 $\times$ H100 GPUs.
The experiments take approximately 1.5 hours for Mistral-base and around 4 hours for Llama-3-Instruct.

\smallskip
\noindent\textbf{Use of AI assistants.}
We used ChatGPT solely for revising the language of the paper.
Note that the revision is exclusively for enhancing the clarity and readability of the text, and not for any other purposes.
\end{document}